\pdfoutput=1

\documentclass[11pt, twocolumn]{article}

\usepackage[preprint]{acl}

\makeatletter
\PassOptionsToPackage{dvipsnames,table}{xcolor}

\makeatother
\usepackage{amsmath}
\usepackage{amssymb}
\usepackage{float}

\usepackage{times}
\usepackage{latexsym}

\usepackage[T1]{fontenc}

\usepackage[utf8]{inputenc}

\usepackage{microtype}

\usepackage{inconsolata}

\usepackage{graphicx}

\usepackage{subcaption}
\usepackage{lipsum}

\usepackage{multirow} 
\usepackage{booktabs}

\usepackage{listings}
\definecolor{mylightorange}{RGB}{255, 200, 100}
\definecolor{myvsgreen}{RGB}{0, 150, 50}
\definecolor{mypurple}{RGB}{128, 0, 128}

\usepackage[noline, ruled]{algorithm2e}

\title{CCQ: Convolutional Code for Extreme Low-bit Quantization in LLMs}

\author{
 \textbf{Zhaojing Zhou},
 \textbf{Xunchao Li},
 \textbf{Minghao Li},
 \textbf{Handi Zhang},
 \textbf{Haoshuang Wang},
 \\
 \textbf{Wenbin Chang},
 \textbf{Yiqun Liu},
 \textbf{Qingqing Dang},
 \textbf{Dianhai Yu},
 \textbf{Yanjun Ma},
 \textbf{Haifeng Wang}
\\
\\
\\
 \textsuperscript{}Baidu Inc.
\\
}

\begin{document}
\maketitle
\begin{abstract}
The rapid scaling of Large Language Models (LLMs) elevates inference costs and compounds substantial deployment barriers. While quantization to 8 or 4 bits mitigates this, sub-3-bit methods face severe accuracy, scalability, and efficiency degradation. 
We propose Convolutional Code Quantization(CCQ), an inference-optimized quantization approach compressing LLMs to 2.0–2.75 bits with minimal accuracy loss. 
Departing from error-prone scalar quantization or slow vector quantization, CCQ integrates a hardware-aware bit-shift encoding and decoding solution with Convolutional Code, Hybrid Encoding, and Code Cluster,  jointly overcoming accuracy-speed bottlenecks. 
We construct a lookup-free encoding space, enabling a linear mapping between the codebook and weight vectors, thereby optimizing inference performance. Meanwhile, by drawing on the concept of data mapping from vector quantization, we minimize the performance degradation of the model under extremely low-bit conditions. 
Experiments demonstrate that CCQ achieves outstanding performance on LLMs across various benchmarks. We compress DeepSeek-V3(671B total parameters) to 184GB and ERNIE-4.5-300B-A47B to 89GB, enabling single-GPU deployment of ERNIE 4.5 and eliminating inter-card communication. 
The 2-bit ERNIE-4.5-300B-A47B model and inference engine have been open-sourced.
\end{abstract}

\section{Introduction}
Large Language Models(LLMs) demonstrate state-of-the-art capabilities across artificial intelligence domains, spanning natural language processing to multi-modal understanding. 
However, these performance gains coincide with significant inference degradation, impeding cost-effective large-scale deployment. 
Contemporary flagship models exemplify this scalability challenge: Qwen2~\cite{qwen2}, Qwen2.5~\cite{qwen2.5}, and Llama 3~\cite{grattafiori2024llama3herdmodels} series have around 100 billion parameters within their most powerful members. 
The emergence of Mixture-of-Experts (MoE) architecture further escalates model scale—evidenced by Llama 3.1 (405B)~\cite{grattafiori2024llama3herdmodels}, DeepSeek-V3 (671B)~\cite{deepseekai2025deepseekv3technicalreport}, Qwen3 (235B)~\cite{qwen3}, and ERNIE 4.5 (300B / 424B) ~\cite{ernie2025technicalreport}. 
Such massive parameterization exacerbates inference latency, primarily due to memory constraints and cross-device communication overhead in distributed environments.

Quantization~\cite{gray1998quantization}, a cornerstone technique for model compression and efficient deployment, reduces parameter precision from 16 bits to 8 or 4 bits, yielding more than 50\% memory savings. 
Common quantization configurations include Weight-Only Quantization(WOQ)~\cite{frantar2022gptq, lin2024awq}, Weight-Activation Quantization(WAQ)~\cite{xiao2023smoothquant} and KV Cache Quantization(KVQ)~\cite{liu2024kivi, hooper2024kvquant}. 
WOQ and KVQ primarily alleviate memory bottlenecks—critical for models with massive parameter counts or long context windows—by compressing weights and key-value caches. 
WAQ, conversely, exploits hardware acceleration for low-precision matrix operations (e.g., INT8/FP8) to achieve compute-bound speedups. 

Recent efforts pursue extremely low-bit quantization (e.g., 2 bits) to further reduce LLM deployment costs. 
However, these methods invariably compromise either model accuracy or inference efficiency, limiting their practical utility.
Post-training quantization (PTQ) algorithms like GPTQ~\cite{frantar2022gptq} and AWQ~\cite{lin2024awq} use scalar quantization to convert weights to low-bit integers. While effective at 4 bits by smoothing outliers, they typically fail at 2 bits due to the constrained numerical space. 

In contrast, vector quantization~\cite{vanbaalen-gptvq, liu2024vptq, tseng2024quip, malinovskii2024pvtuning} aims for higher compression and better performance by representing weights as indices into a finite codebook of high-precision vectors. During quantization, weights are clustered and replaced by codebook indices; storage is compressed by saving these indices and the codebook. Inference reconstructs weights via codebook lookup.

Despite offering superior compression fidelity over scalar quantization, vector quantization introduces critical performance bottlenecks. The dequantization process requires large-scale, non-contiguous codebook lookups based on indices. This irregular memory access pattern violates data locality, prevents efficient hardware utilization, causes pipeline stalls, and introduces latency proportional to codebook size, severely degrading LLM inference throughput.

To circumvent the performance drawbacks associated with inference-unfriendly codebooks while preserving the benefits of mapping high-dimensional vectors to low-dimensional vectors in vector quantization, we propose convolutional code quantization by integrating convolutional encoding with scalar quantization. By combining innovative hybrid encoding and code cluster, we achieve several extremely low-bit quantization configurations.
Initially, we construct a codebook based on convolutional codes, leveraging their inherent properties to achieve efficient bitshift decoding. Subsequently, we employ scalar quantization to establish a linear mapping between high-dimensional and low-dimensional vectors. 
This approach not only obviates the need for storing a codebook but also replaces time-consuming indexing operations with linear computations, thereby offering both performance and efficacy benefits. Our main contributions are as below: 

\begin{itemize}
    \item \textbf{Convolutional Code Quantization (CCQ):} We propose a novel weight-only PTQ solution for extreme compression of LLMs. By integrating convolutional coding theory, hybrid encoding, and code cluster algorithms, CCQ achieves around 70\% model size savings at 2-bit precision compared with 8-bit counterpart, while maintaining near-lossless accuracy on both DeepSeek-V3 and ERNIE 4.5 models.
    
    \item \textbf{Bit-Shift Decoding Inference:} Our co-designed bit-shift decoding strategy establishes inference latency similar to scalar-based quantization. This enables unprecedented deployment of ERNIE-4.5-300B-A47B (the largest ERNIE 4.5 language model) on a single NVIDIA H20 GPU without intra-node communication overhead.
    
    \item \textbf{Open-Source:} We publicly release 2-bit quantized ERNIE 4.5 variants\footnotemark with around 2\% accuracy degradation relative to baselines. The inference code is available at \url{https://github.com/PaddlePaddle/FastDeploy/tree/develop}.
\footnotetext[1]{https://huggingface.co/baidu/ERNIE-4.5-300B-A47B-2Bits-Paddle}
\end{itemize}

\section{Background and Related Work}
\subsection{Scalar-based Quantization}
RTN (Round-to-Nearest) is the most straightforward method for quantizing weight and activations by using the min-max values calibrated from a dataset. 
However, the quantization error increases significantly when the number of bits is small, especially when significant outliers exist. 
GPTQ~\cite{frantar2022gptq} updates the weight values based on the previous quantization error and the inverse of the Hessian matrix calculated from the activations. 
AWQ~\cite{lin2024awq} transferred outliers from weights to activations to smooth the distributions. Group-wise quantization is applied to both methods. 
However, these methods still suffer from non-negligible accuracy degradation even under 3/4-bit quantization, primarily constrained by the limited dynamic range of integer representation.

SqueezeLLM~\cite{kim2023squeezellm} implements non-uniform quantization and Hessian Metrics, GPTAQ~\cite{li2025gptaq} explicitly minimizes the quantization error as well as the accumulated asymmetry error for better performance under low-bit quantization. 
HQQ~\cite{badri2023hqq} and HQQ+~\cite{badri2023hqq+} introduce Hyper-Laplacian distribution to better capture the heavy-tailed outlier errors compared to MSE. 
However, substantial performance degradation still exists for 2 bits and even lower bits. 

To improve the extremely low-bit model performance, Quantization-Aware-Training (QAT) is introduced in recent works. LLM-QAT~\cite{liu2023llm} achieves 4 bits on weights and KV Cache. But distillation is required to maintain the accuracy, which leads to heavy training pipelines. BitNet~\cite{wang2023bitnet} achieves 1.58-bit quantization by introducing QAT from the pre-training phase. However, the prolonged training duration and inherently limited scalability significantly constrain their practical deployment. OneBit ~\cite{xu2024onebitextremelylowbitlarge} and LittleBit ~\cite{lee2025littlebitultralowbitquantization} attempt to combine matrix decomposition and QAT for higher compression ratios. However, they still fail to eliminate the quantization cost and training unstablility. 

\subsection{Vector-based Quantization}
To overcome the performance limit of scalar-based quantization, vector-based quantization methods are introduced for low-bit quantization. 

Vector Quantization~\cite{gray1984vector} (VQ) is originally designed for lossy data compression. 
It works by categorizing a large set of vectors into a small one based on some clustering algorithms, like K-means. 
In the LLMs, index maps and codebooks are stored as the compressed weights to achieve the extreme compression ratio.

The main problem for VQ is how to balance the accuracy and its quantization cost. 
AQLM~\cite{egiazarian2024extreme, malinovskii2024pvtuning} attempts to maximize the codebook's dimensionality, leaving the unstructured indexing the main bottleneck for inference speed. 
QUIP\#~\cite{tseng2024quip} uses incoherence processing and E8 lattice for smaller codebook size, but it induces post-processing for Random Hadamard Transform (RHT), which hurts inference speed as well. 
VPTQ~\cite{liu2024vptq} parallelly updates the vector in one column to avoid error accumulation and introduces residual quantizations for higher accuracy. However, it also falls short in improving inference performance.  

\subsection{Convolutional Code}
Convolutional codes represent a classic technique in channel coding, which involves continuous encoding of each state within the channel. 
Their primary characteristic lies in preserving the memory effect of the channel, where each current state's code value is derived from the transition of the preceding state. 

Fischer et al.~\cite{marcellin1990trellis} applied a representation of convolutional codes known as trellis codes to quantization, thereby introducing Trellis-Coded Quantization (TCQ). 
Subsequently, QTIP~\cite{tseng2024qtip} further enhances TCQ in terms of memory usage and decoding speed. 
Firstly, QTIP employs incoherent processing to transform weights into an independent Gaussian-like source, which is well-suited for the trellis code. 
Secondly, the researchers devised computation-based pseudo-Gaussian codes to circumvent the need for storing codebooks and defining network architectures. 
Despite achieving favorable outcomes, QTIP still suffers from the drawback of computational complexity.




\section{Methodology}
This section presents the core elements of our proposed method. 
Firstly, in \hyperref[sec:ccq]{Section 3.1}, we will elaborate on how we integrate convolutional codes with scalar quantization to construct a codebook capable of achieving linear mapping between high-dimensional and low-dimensional vectors, along with an overview of the framework of our algorithms. 
Subsequently, in \hyperref[sec:he]{Section 3.2}, we demonstrate how we flexibly combine different convolutional codes to attain a higher compression ratio through hybrid encoding.
Furthermore, in \hyperref[sec:ec]{Section 3.3}, we transcend the limitations imposed by data types in encoding and storage by employing the code cluster, thereby achieving significantly superior compression performance. 
In \hyperref[sec:scale-compress]{Section 3.4}, we introduce the implementation details of scale compression and optimization. 
In \hyperref[sec:infer]{Section 3.5}, we present the performance optimization in the inference implementation.

\subsection{Convolutional Code Quantization}
\label{sec:ccq}
\begin{figure}[t]
    \centering
    \includegraphics[width=1\linewidth]{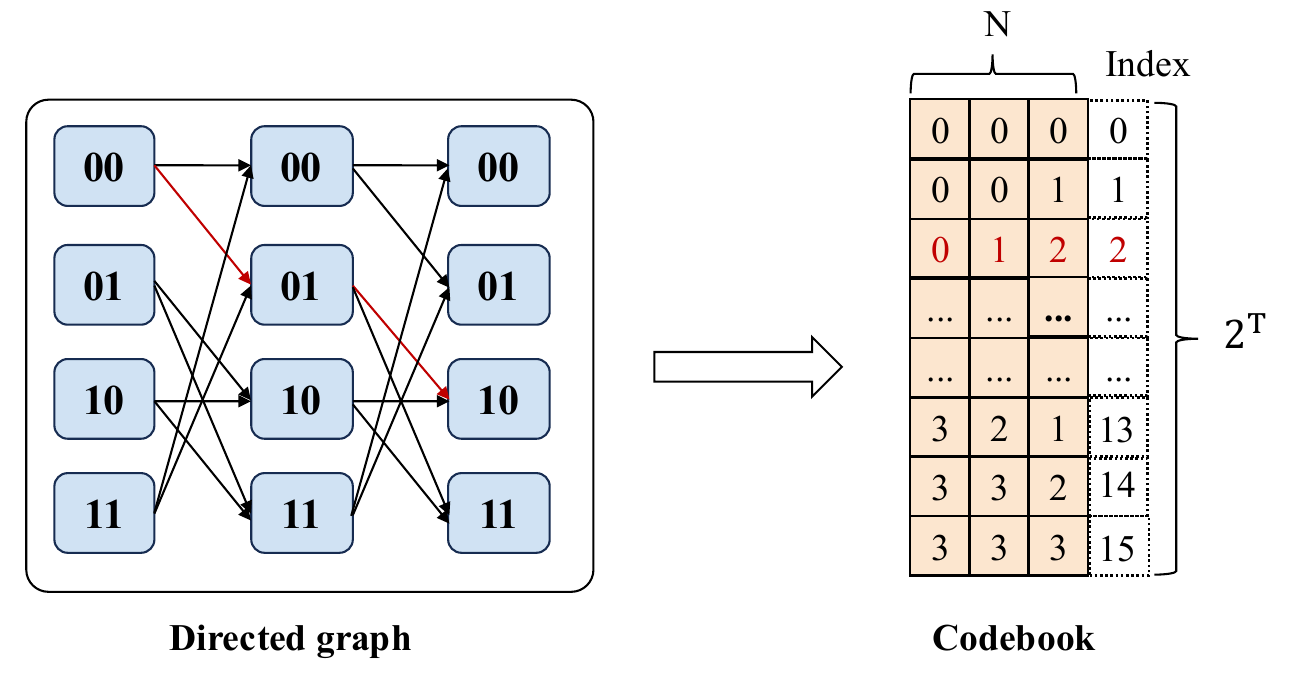}
    \caption{Convolutional code codebook construction. Taking (L=2, N=3, S=1) configuration as an example, all the connection relationships among states in the directed graph are consolidated into a codebook. In the codebook, each row represents N consecutive states, and the corresponding convolutional code is the index.}
    \label{fig:codebook}
\end{figure}

\begin{figure*}[ht]
    \centering
    \includegraphics[width=\textwidth]{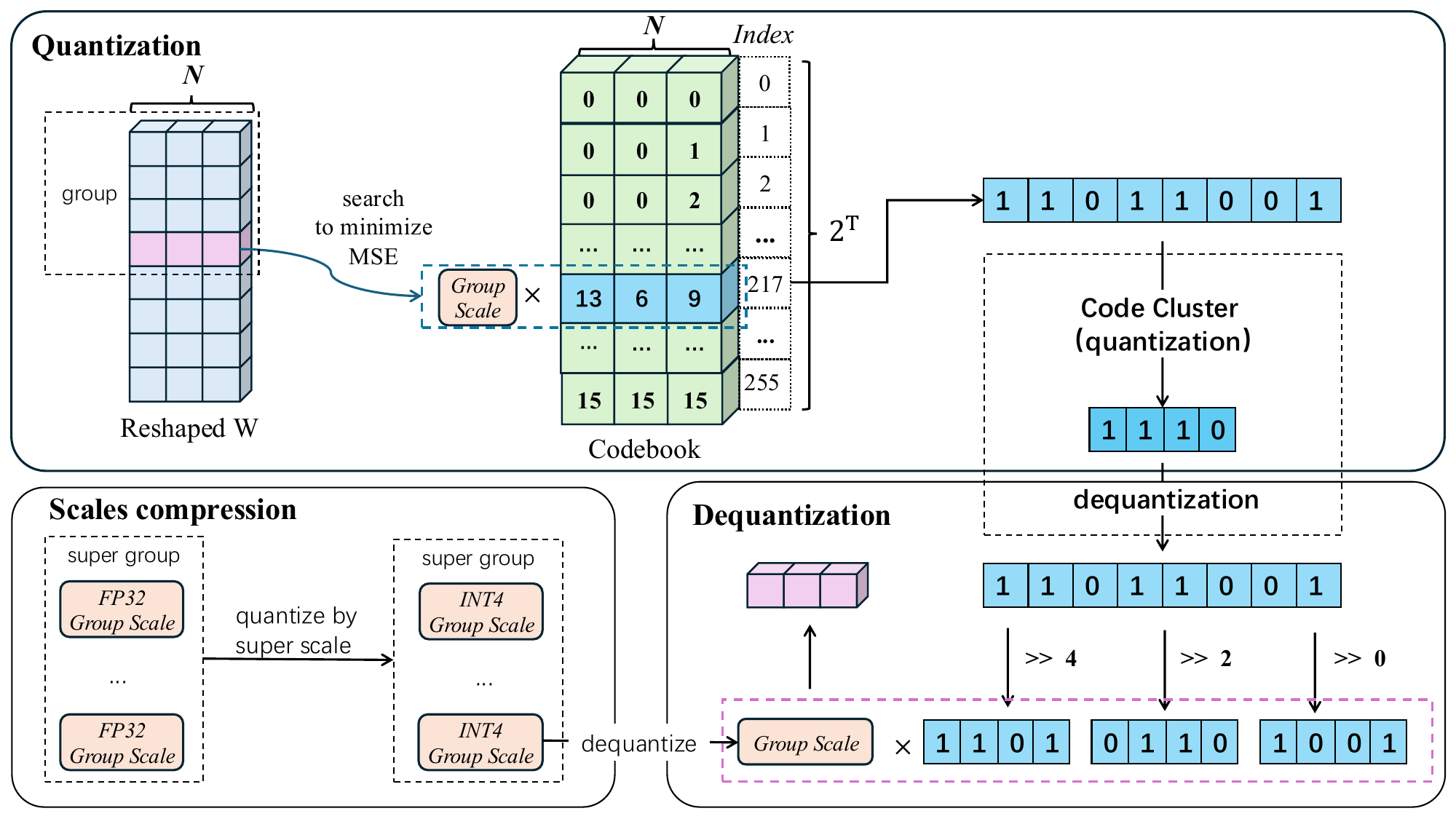}
    \caption{A schematic diagram of the CCQ. The dequantization part only involves shifting and dequantization calculations, and does not require the storage of a codebook. For the sake of simplicity, the codebook depicted in the figure is an example with parameters (L=4, N=3, S=2). In reality, as mentioned in \hyperref[sec:ec]{Section 3.3}, we perform the code cluster under the configuration of (L=6, N=4, S=3). The figure merely serves as an illustrative example.}
    \label{fig:encode}
\end{figure*}

Unlike TCQ and QTIP, which perform sequential encoding on long sequences, we employ a vector-quantization-like approach to compress weights into vector groups. 
By simplifying the design of the codebook, we aim to reduce computational complexity during the encoding process. 
Consequently, we redefine convolutional codes from a directed graph representation to a finite set of vectors (i.e., the codebook).

\textbf{Definition of Convolutional Code Codebook}. We define a convolutional encoding configuration as the triplet (L, N, S), where L denotes the number of bits required for each state, N represents the number of states involved in a single encoding operation, and S indicates the number of transition bits between adjacent states. 
Here we define transition bits as the bits that do not overlap with the previous state. 
Figure \ref{fig:codebook} depicts all possible transition forms between adjacent states for the configuration (L=2, N=3, S=1). 
Each state requires $L=2$ bits for representation. 
The transition between adjacent states is achieved using $S=1$ transfer bit to update the state. 
Consequently, there are $L - S=1$ redundant bits between adjacent states. 
Given the overlapping bits between adjacent states, after applying convolutional coding, we only need to store the code symbols of the initial state along with the state transition bits to fully represent the entire state sequence. 
For instance, considering the three states connected by the red line in Figure \ref{fig:codebook} ($00\rightarrow 01\rightarrow10$), the final stored code would be $0010$. 
These three state values can then be placed in the index 2 (0010) of the codebook. Following this approach, we successfully transform the directed graph into a code table comprising a collection of vectors, each of length N.

Evidently, for a given encoding configuration, the total number of bits in a set of convolutional codes is calculated as $T = L + (N-1)*S$. Subsequently, we can generate a corresponding codebook $\mathbf{C} \in \mathbb{Z}^{2^T\times N}$.

Taking 4-bit quantization as an example in weight quantization, where each quantized weight requires 4 bits for storage, in the case of convolutional encoding with parameters set as (L=4, N=3, S=2), we can store three quantized values using only 8 bits, achieving an equivalent compression effect of 2.66 bits per weight (bpw). 
Compared to the 4-bit weight quantization, this results in a 33\% improvement in compression rate while maintaining the equivalent numerical precision as 4-bit quantization. 

\textbf{Bitshift Decoding}. Due to the repetitive nature between adjacent states in convolutional codes, we can rapidly decode convolutional codes through bit shifting. 
Consequently, within the framework of convolutional code quantization, the dequantization process only involves bit-shifting operations and dequantization calculations for scalar quantization, which undoubtedly facilitates inference implementation more than index lookup in vector quantization. 
The specific decoding process can be referred to in Figure \ref{fig:encode}.

\textbf{Group-wise Quantization}. Although the memory characteristic of convolutional codes offers convenience in storage and decoding, it also restricts the non-coherence of quantized weights. 
When weights contain a significant number of outliers, the coherence of quantized values diminishes, which is detrimental to convolutional coding. 
Therefore, we introduce group-wise quantization to mitigate the impact of outliers. 
The quantization granularity at the group-wise level not only ensures the quantization precision of scalar quantization but also reduces the encoding loss during convolutional coding. In scenarios where the group size is not a multiple of N, we will pad each group with zeros to fulfill the encoding requirements. 
This approach will not compromise the accuracy of quantization.

\textbf{The Overall Quantization Algorithm}. The complete process of the convolutional code quantization method is demonstrated in Algorithm \ref{algo:whole algo}.
Firstly, we group the weights and calculate the quantization parameters for each group. 
Subsequently, we generate the encoding space based on the convolutional code configuration (at this stage, the codebook contains numerical values within the integer domain, denoted as $\mathbf{C} \in \mathbb{Z}^{2^T\times N}$). 
We then transform these values into the same floating-point domain as the original weights using the group scale (resulting in $\mathbf{C} \in \mathbb{R}^{2^T\times N}$). 
Finally, we compute the Mean Squared Error (MSE) loss between each weight and every vector in the codebook. 
The optimal code value is determined by searching for the one that yields the minimum MSE loss. 

It is important to note that there is no need to store the actual vector values obtained from the search—only their corresponding indices need to be retained. 
The step is consistent with vector quantization, but the key difference lies in the fact that our method does not require storing a large codebook to maintain the mapping between indices and vectors.

In our convolutional encoding quantization scheme, the indices obtained from the search directly represent the final convolutional code values. 
During subsequent dequantization, the true quantized values can be retrieved through bit-shifting operations, and the original weights can be reconstructed via dequantization computations.

\begin{algorithm}
\footnotesize
\SetAlgoLined
\SetKwInOut{Input}{Input} 
\SetKwInOut{Output}{Output} 
\Input{
    $\mathbf{W} \in \mathbb{R}^{d_{o} \times d_{i}}$: Original Weight \\
    $g$: group size \\
    $(L, N, S)$: Encoding configuration 
}
\Output{
    $\mathbf{W_q} \in \mathbb{Z}$: Weight with Convolutional Code
}
\BlankLine
$\mathbf{W_g} \gets$ Reshape $\mathbf{W}$  \tcp*{$\mathbf{W_g} \in \mathbb{R}^{\frac{d_o \times d_i}{g}\times \frac{g}{N}\times N}$}

${group\ scale} \gets \mathbf{W_g}$ \tcp*{$group\ scale \in \mathbb{R}^{\frac{d_o \times d_i}{g}}$}

$\mathbf{C}$ = Construct\_codebook$(L, N, S)$ \tcp*{$\mathbf{C} \in \mathbb{Z}^{2^T\times N}$}

$\mathbf{C}$ = $\mathbf{C} \times {group\ scale}$ \tcp*{$\mathbf{C} \in \mathbb{R}^{\frac{d_o \times d_i}{g} \times 2^T\times N}$}

$Indices \gets \underset{i\in 2^T}{\arg\min}\|\boldsymbol{v}-\mathcal{C}_i\|^2,\forall\boldsymbol{v}\in\mathbf{W_g}$

$\mathbf{W_q} \gets$ Reshape $Indices$

\caption{Convolutional Code
Quantization}
\label{algo:whole algo}
\end{algorithm}

\subsection{Hybrid Encoding}
\label{sec:he}
Although our method can achieve arbitrary compression rates by adjusting the encoding configuration, the actual storage compression effect can only be realized when the code aligns with practical data types. 
Specifically, the storage bits for convolutional coding must be exact multiples of 8 to be stored using data types such as INT8 or INT16. 
However, achieving a high compression rate typically necessitates an increase in the number of encoding states. 
Due to the inherent correlation in convolutional coding, an excessive number of encoding states can significantly exacerbate precision loss. 
Therefore, to attain a higher compression rate, we propose the concept of hybrid encoding. 
Within a single group, multiple distinct encoding configurations are employed in a hybrid manner to meet storage requirements. 
Through meticulous design, we successfully alternate between the (L=3, N=3, S=2) and (L=3, N=4, S=2) configurations, encoding seven numerical values into 16 bits, ultimately achieving an equivalent bit rate of 2.28 bpw.

\subsection{Code Cluster}
\label{sec:ec}
By employing hybrid encoding, we can alleviate the constraints imposed by storage data types, but achieving further compression remains challenging. 
To realize true 2-bit weight quantization, we innovatively propose the code cluster algorithm that entirely breaks free from the limitations of encoding configurations on compression rates.

Our convolutional code quantization is conducted at a group-wise granularity. 
Given the vast encoding space, the encoding values utilized on each channel are quite limited. 
This observation raises the question: can we retain only the frequently used encoding values and eliminate the redundant ones, thereby enhancing the overall compression rate by reducing the encoding space? 
Naturally, the idea of clustering comes to mind to distill the significant encoding values. However, clustering is a nonlinear operation, and its data reconstruction process inevitably requires index lookups.

\begin{figure}[h!]
    \centering
    \includegraphics[width=0.45\textwidth]{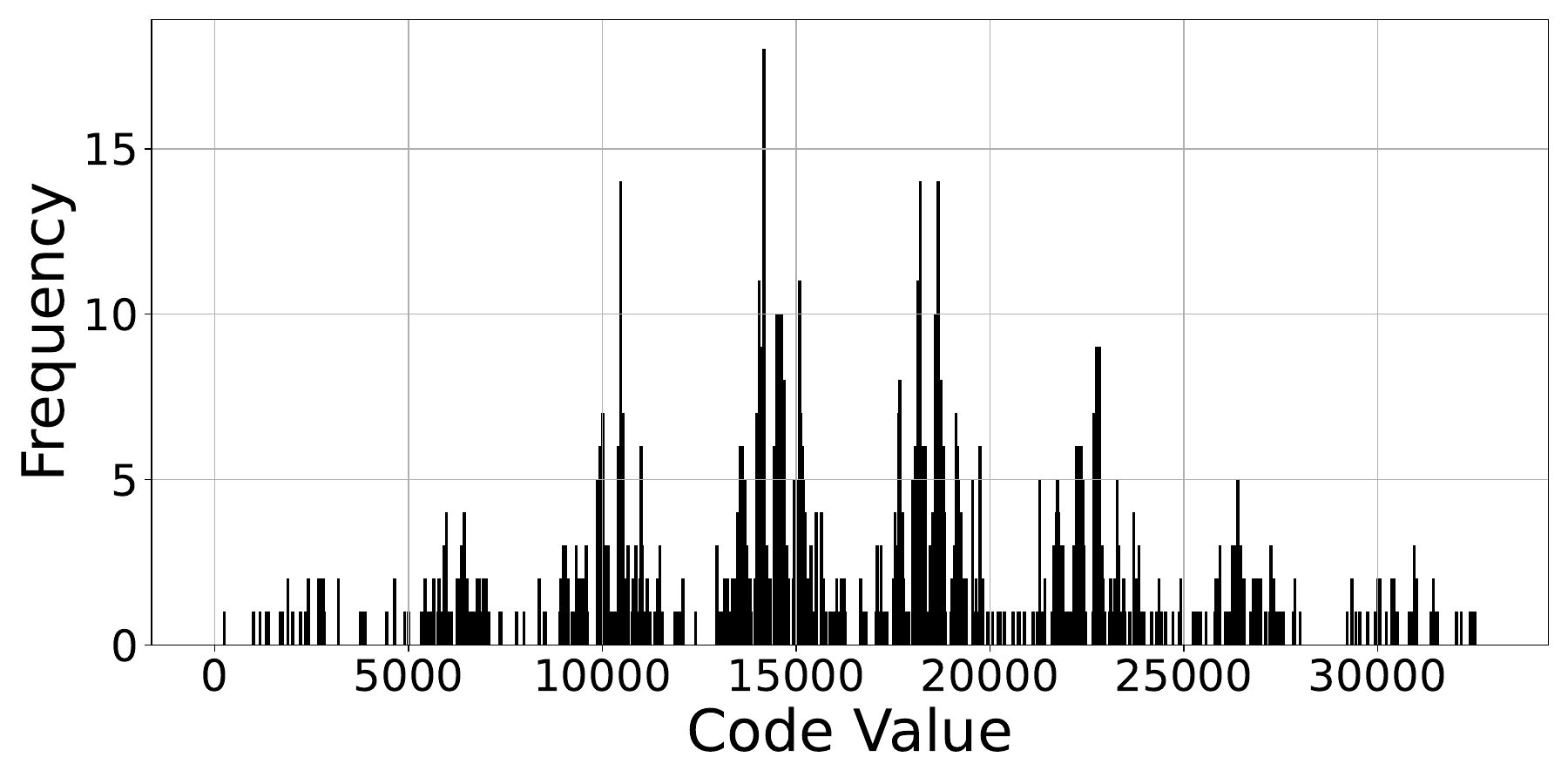}
    \caption{The code value distribution diagram of convolutional coding quantization on an output channel under the configuration of (L=6, N=4, N=3).}
    \label{fig:code distribution}
\end{figure}

Upon examining the numerical distribution of convolutional encoding quantization (Figure \ref{fig:code distribution}), we observe that the convolutional codes on each channel approximately follow a normal distribution, which is suitable for uniform quantization. 
Consequently, we propose utilizing uniform quantization to accomplish the code cluster, mapping any arbitrary encoding space into an 8-bit numerical range. 
As a result, all data reconstruction operations can be accomplished through computation. 
The computational formula for the code cluster is presented as follows:

\begin{align}
\mathbf{C'} &= \mathcal{Q}(\mathbf{C})=\left  \lfloor \frac{\mathbf{C}-\beta }{\alpha}  \right \rceil
\end{align}
where $\mathbf{C} \in [0, 2^T-1]$, $\mathbf{C'} \in [0, 255]$, $\alpha$ means code scale and $\beta$ means code zero-point in asymmetric quantization.

Specifically, we first perform convolutional code quantization with parameters (L=6, N=4, S=3) on the weights of each group. 
Subsequently, we conduct clusters of the code values along the dimension of output channels and map them to an 8-bit numerical range. 
This approach enables the storage of four parameters using only UINT8, achieving an equivalent compression effect of 2 bpw.

At this point, although the code cluster restricts the encoding space, the true quantized value of the weights is 6 bits. 
When combined with the group-wise quantization granularity, it can effectively ensure the numerical precision of critical weights. Consequently, the impact on the model's performance is negligible.

\subsection{Scale Optimization and Compression}\label{sec:scale-compress}
\textbf{Scale Optimization}: After determining the encoded values through coding search, we can compute the quantization loss by dequantizing these values and comparing them with the original weights. 
The quantization error formula for a group of weight is as follows:

\begin{equation}
\begin{split}
Error &=  {\textstyle \sum_{i=0}^{g-1}} \left \| \mathbf{W}_i -\mathcal {DQ} (\mathbf{W}{^q_i}) \right \| ^2 \\
&= {\textstyle \sum_{i=0}^{g-1}}\left \| \mathbf{W}_i - \mathbf{W}{^q_i}\cdot S \right \| ^2
\end{split}
\end{equation}

where $\mathbf{W}$ represents the original weight of a group, $\mathbf{W_q}$ denotes the corresponding quantized values, $g$ denotes the group size, $\mathcal {DQ}$ signifies the dequantization function, and $S$ stands for the scale. 
For the sake of simplifying calculations, we directly use the decoded values to compute the quantization error, as bitshift decoding does not affect the accuracy of the quantized values.

At this stage, the scale $S$ can be regarded as the sole variable. We can further reduce the quantization loss by updating the scale. 

\begin{equation}
\begin{split}
\frac{\mathrm{d}Error}{\mathrm{d}S}&= 2\sum_{i=0}^{g}(\mathbf{W}{^q_i})^2S - 2\sum_{i=0}^{g}(\mathbf{W}{_i}\mathbf{W}{^q_i})
\end{split}
\end{equation}

By taking the derivative of the scale and setting $\frac{\mathrm{d}Error}{\mathrm{d}S}$ to zero, we can determine the minimum value of the current loss.

\begin{equation}
\begin{split}
S=\frac{ {\textstyle \sum_{i=0}^{g-1}}\mathbf{W\cdot W}^q_i}{{\textstyle \sum_{i=0}^{g-1}}(\mathbf{W}{^q_i})^2}
\end{split}
\end{equation}

\textbf{Scale Compression}: In the group-wise quantization, each group has its own scale factor. 
As the number of groups increases, the storage overhead associated with these scale factors becomes non-negligible. 
To address this, we propose quantizing the scale factors by channel-wise super scales and storing them in a low-bit format. 
Additionally, we observe that for the aforementioned 2.66 bpw (with L=4, N=3, S=2) and the hybrid encoding scheme yielding 2.28 bpw, the group size is not divisible by the number of encoding states $N$. 
Consequently, the last code value in each group introduces redundant bits. 
Taking a group size of 64 as an example, under the (L=4, N=3, S=2) configuration, 22 code values are required to encode 64 parameters. 
In this scenario, only the first 4 bits of the last code value are valid, while the remaining 4 bits are redundant and need to be discarded after decoding. 
We can leverage these 4 redundant bits to store the quantized scale factor, ultimately achieving an equivalent bpw of 2.75. 
Similarly, for the hybrid encoding scheme with 2.28 bpw, 10 code values are necessary to encode 64 parameters, leaving 13 redundant bits that can be utilized to store the quantized scale factor, resulting in an equivalent bpw of 2.5. 
When it comes to the code cluster, redundant bits cannot be utilized, and additional consideration must be given to the storage overhead of the scale. For the code cluster of (L=6, N=4, S=3), the equivalent bit width is 2.06 bpw. The formula for calculating the equivalent bit width is presented below:

\begin{equation}
\begin{split}
bpw = \left\{\begin{matrix} 
\frac{T*{\lceil \frac{g}{N} \rceil }}{g}, & \text{if } g \equiv 1 \pmod{N} \\  
\frac{T*{ \frac{g}{N} }}{g}+\frac{scale\_bits}{g}, & \text{if } g \equiv 0 \pmod{N}
\end{matrix}\right.
\end{split}
\end{equation}
where $T$ represents the total number of bits for a set of convolutional codes. 
However, when the code cluster is used, $T$ denotes the number of bits required for the quantized code values, specifically $T = 8$.

\subsection{Dequantization}
\label{sec:infer}
In this section, we present the detailed implementation strategies for dequantizing weights within individual groups, as outlined in Algo.~\ref{algo:w275-dequant} and Algo.~\ref{algo:w2-dequant}, and summarize the corresponding configurations in Table~\ref{tab:dequant-config}. 
\begin{table}[h]
\renewcommand{\arraystretch}{1.2}
\centering
\caption{Dequantization configuration for different encoding strategies.}
\label{tab:dequant-config}
\resizebox{\linewidth}{!}{
\begin{tabular}{@{}ccccccc@{}}
\toprule
(L,N,S) & bpw & Weight Mask & Scale Mask & Weight Bit-Shift & Code Cluster \\
\midrule
(4,3,2) & 2.75 & 0xF & 0xF & [4,2,0] & No \\
(3,3,2)(3,4,2) & 2.5 & 0x7 & 0x1FFF & [13,11,9,6,4,2,0] & No \\
(6,4,3) & 2.06 & 0x3F & 0xF & [9,6,3,0] & Yes \\
\bottomrule
\end{tabular}
}
\end{table}
For the standard dequantization of Hybrid Encoding, we use Algo.~\ref{algo:w275-dequant} to dequant weights. 
For example, the (L=4, N=3, S=2) configuration, which is 2.75 bpw, we set weight bit-shift $\mathbf{W_{shift}}$ to [4, 2, 0] with 0xF as weight mask $\mathbf{w_{mask}}$; For the hybrid encoding configuration (L1=3, N1=3, S1=2, L2=3, N2=4, S2=2), we pass $\mathbf{W_{shift}}$ with [13, 11, 9, 6, 4, 2, 0] and weight mask $\mathbf{w_{mask}}$ with 0x7. 
As discussed in Section~\ref{sec:scale-compress}, we leverage the redundant bits within each weight group to store additional information. 
During the dequantization process for each computational block, the algorithm efficiently extracts the scaling factors (scale) that are embedded within these redundant bits of the loaded weight matrix $\mathbf{W_q}$.

In contrast to the Hybrid Encoding, the encoding configuration (L=6, N=4, S=3) with Code Cluster employs $\mathbf{W_{shift}}$ of [9, 6, 3, 0] and $\mathbf{w_{mask}}$ of 0x3F, as detailed in Algo.~\ref{algo:w2-dequant}. 
This configuration necessitates two distinct dequantization operations within the main computation loop. 
The first dequantization operation expands the quantized weights from UINT8 to UINT16, thereby enlarging the numerical space for subsequent processing. 
The second operation applies different bit-shift operations to dequantize the weights, following a similar approach to Algo.~\ref{algo:w275-dequant}.
Another key distinction from Algo.~\ref{algo:w275-dequant} stems from the fact that the code cluster eliminates redundant bits entirely, consequently requiring explicit loading of corresponding group-wise quantization scales for each computation block during the main processing loop.

\begin{algorithm}[t]
\footnotesize
\SetAlgoLined
\SetKwInOut{Input}{Input} 
\SetKwInOut{Output}{Output} 
\Input{
    $\mathbf{W_q}$: compressed quantized weights \\
    $\mathbf{s_{super}}$: super scale \\
    $\mathbf{zp}$: weight zero-point \\
    $\mathbf{W_{shift}}$: bit shift of quantized weight\\
    $\mathbf{w_{mask}}$: weight mask \\
    $\mathbf{s_{mask}}$: scale mask \\
}
\Output{
    $\mathbf{W_{dq}}$: dequantized weight
}
\BlankLine  

$\mathbf{p_{w}} \gets$ length of $\mathbf{W_{shift}}$\;

$s_{uint} \gets $ the last value of $\mathbf{W_q}$\;

$\mathbf{s} \gets (s_{uint} \ \& \ \mathbf{s_{mask}}) \times \mathbf{s_{super}}$\;

\For{each element $d_w$ in $\mathbf{W_{dq}}$ with index $idx$}{

$q_w \gets \mathbf{W_q}[{idx/\mathbf{p_{w}}}]$\;

$w_{uint} \gets (q_w \gg \mathbf{W_{shift}}[{idx \ \mathbf{mod} \  \mathbf{p_{w}} }]) \ \&\ \mathbf{w_{mask}}$\;  

$d_w \gets (w_{uint} - \mathbf{zp}) \times \mathbf{s}$\;  
}
\caption{Dequantization without Code Cluster}
\label{algo:w275-dequant}
\end{algorithm}  

\begin{algorithm}[t]
\footnotesize
\SetAlgoLined
\SetKwInOut{Input}{Input} 
\SetKwInOut{Output}{Output} 
\Input{
    $\mathbf{W_q}$: compressed quantized weights \\
    $\mathbf{s_q}$: quantized scale \\
    $\mathbf{s_{super}}$: super scale \\
    $\mathbf{zp}$: weight zero-point \\
    $\mathbf{s_{code}}$: code scale \\
    $\mathbf{zp_{code}}$: code zero-point \\
    $\mathbf{W_{shift}}$: bit shift of quantized weight\\
    $\mathbf{s_{shift}}$: bit shift of quantized scale\\
    $\mathbf{w_{mask}}$: weight mask \\
    $\mathbf{s_{mask}}$: scale mask \\
}
\Output{
    $\mathbf{W_{dq}}$: dequantized weight
}
\BlankLine  

$\mathbf{s} \gets ((\mathbf{s_q} \gg \mathbf{s_{shift}})\ \& \ \mathbf{s_{mask}}) \times \mathbf{s_{super}}$\;

$\mathbf{p_{w}} \gets$ length of $\mathbf{W_{shift}}$\;

\For{each element $d_w$ in $\mathbf{W_{dq}}$ with index $idx$}{

$q_w \gets \mathbf{W_q}[{idx/\mathbf{p_{w}}}]$\;  

$w_{uint} \gets \mathbf{round}(q_w \times \mathbf{s_{code}} + \mathbf{zp_{code}})$\;  

$w_{uint} \gets (w_{uint} \gg \mathbf{W_{shift}}[{idx \ \mathbf{mod} \  \mathbf{p_{w}} }]) \& \ \mathbf{w_{mask}}$\;  

$d_w \gets (w_{uint} - \mathbf{zp}) \times \mathbf{s}$\;  
}
\caption{Dequantization with Code Cluster}
\label{algo:w2-dequant}
\end{algorithm}     

\section{Experiments}
\subsection{Settings}

\paragraph{Models and Datasets.} We conduct rigorous evaluations using 2 bpw, 2.5 bpw, and 2.75 bpw on two state-of-the-art large language models: DeepSeek-V3-0324 and ERNIE-4.5-300B-A47B. 
%
The experiments compare CCQ against 8-bit and 4-bit weight-only quantized models.


We evaluate our model under a unified evaluation protocol across several key benchmarks:
\begin{itemize}
    \item \textbf{Zero-shot} performance on mathematical reasoning (GSM8K~\cite{cobbe2021training}), Chinese mathematics (CMath~\cite{wei2023cmath}), and multi-step reasoning (MUSR~\cite{sprague2023musr}).
    \item \textbf{3-shot} performance on complex task handling (BBH~\cite{srivastava2022beyond}) and reading comprehension (DROP~\cite{dua2019drop}).
    \item \textbf{5-shot} performance on Chinese comprehensive evaluation (C-Eval~\cite{huang2023ceval}), and massive multi-task language understanding (MMLU~\cite{hendryckstest2021}).
\end{itemize}
These datasets span critical capabilities including mathematics, reasoning, question answering, reading comprehension, and general tasks. All evaluations use 
 the generation configuration, as per the model's recommended settings.


\paragraph{Baselines.} We benchmark our method against 8-bit and 4-bit weight-only quantization. 
For DeepSeek-V3~\cite{liu2024deepseek}, comparisons are conducted using open-source quantization models from Hugging Face. 
For ERNIE 4.5, we adopt Weight-only INT8 (WINT8) as the baseline.

All results across tasks and algorithms are generated under identical evaluation rules.
%
A comprehensive comparison with the aforementioned weight-only quantization approaches is provided in \hyperref[sec:mr]{Section 4.2}.

\subsection{Main Results}
\label{sec:mr}



\paragraph{Results on DeepSeek-V3.} As summarized in Table~\ref{tab:dsv3}, our proposed \textbf{CCQ} method achieves comparable performance with mainstream 8-bit and 4-bit weight-only quantization techniques. 
%

%


%
%
%
%
While 2-bit RTN weight-only quantization induces a performance cliff with catastrophic accuracy collapse, our 2.06-bit CCQ method reduces memory usage by up to 71.3\% from 8-bit models with marginal performance degradation. 
Section~\ref{sec:infer} provides further hardware-aware optimization analyses.

\begin{table}[h]
\renewcommand{\arraystretch}{1.2}
\centering
\caption{Weight-only quantization results of DeepSeek-V3-0324.}
\label{tab:dsv3}
\resizebox{\linewidth}{!}{
\begin{tabular}{@{}ccccccc@{}}
\toprule
Model                    & \multirow{2}{*}{bpw} & \multicolumn{4}{c}{DeepSeek-v3-0324}                              \\
Dataset                  &                      & Memory & GSM8K          & C-Eval         & MMLU & Avg         \\ \midrule
Baseline                 & 8                    & 642GB  & 96.21          & 87.31          & 86.50 & 90.01         \\
GPTQ\footnotemark                     & 4                    & 346GB  & 94.77          & 87.66          & 86.60 & 90.04          \\
AWQ\footnotemark                      & 4                    & 328GB  & 96.13          & 87.22          & 86.38 & 89.43          \\
\multirow{3}{*}\textbf{CCQ (our)} & 2.75         & 230GB  & 95.68 & 85.08 & 85.24 & 88.67 \\
                         & 2.5                  & 212GB  & 95.60         & 84.11          & 85.07 & 88.26             \\
                         & 2.06                 & \textbf{184GB}  & 94.92          & 84.07          & 84.29 & 87.76         \\ \bottomrule
\end{tabular}
}
\end{table}
\footnotetext[2]{GPTQ:%
\url{https://huggingface.co/ISTA-DASLab/DeepSeek-V3-0324-GPTQ-4b-128g-experts}}
\footnotetext[3]{AWQ:%
\url{https://huggingface.co/cognitivecomputations/DeepSeek-V3-0324-AWQ}}

\paragraph{Results on ERNIE 4.5.} As quantified in Table~\ref{tab:ernie-4.5}, our 2 bpw quantized model reduces GPU memory consumption by 68.33\% compared to the WINT8 baseline, while constraining accuracy degradation to around 2\%. CCQ establishes a cost-effective inference pathway that enables practical single-GPU deployment on hardware such as NVIDIA H20 devices. 

However, on both DeepSeek-V3 and ERNIE 4.5, accuracy drops of CCQ primarily occur in 5-shot C-Eval and MMLU tasks due to instruction-following failures. Future work should focus on improving model adherence to instructions.

\begin{table*}[h]
\renewcommand{\arraystretch}{1}
\centering
\caption{Weight-only quantization results of ERNIE 4.5.}
\label{tab:ernie-4.5}
\resizebox{\linewidth}{!}{

\begin{tabular}{cccccccccc}
\hline
Model   & \multicolumn{9}{c}{ERNIE-4.5-300B-A47B}                                        \\
Dataset & Memory & GSM8K  & CMath & MUSR & BBH & DROP & C-Eval   & MMLU & Avg   \\ \hline
WINT8   & 281GB & 96.6 & 96.7 & 69.9 & 94.3 & 91.1 & 90.6 & 86.5 & 89.39   \\
WINT4   & 149GB & 96.21 & 96.5 & 69.67 & 94.43 & 91.17 & 89.84 & 86.16 & 89.14 \\
WINT2   & 89GB & 95.98 & 96.0 & 69.57 & 92.02 & 89.97 & 85.39 & 82.58 & 87.36  \\ \hline
\end{tabular}
}
\end{table*}

\section{Inference Performance}

We conduct comprehensive benchmarks on both individual GEMV (General Matrix-Vector Multiplication) and Grouped-GEMM (Grouped General Matrix-Matrix Multiplication) operators with various matrix shapes and batch size M on the NVIDIA H20 GPU. 
For GEMV, we compare the performance of VPTQ~\cite{liu2024vptq}, standard WINT2, and our proposed CCQ implementations with bpw equal to 2.06. 
%
VPTQ is evaluated using its official codebase with recommended inference configurations(bpw=3), while both standard WINT2 and CCQ are implemented with similar Triton-based kernels~\cite{tillet2019triton} to ensure a fair comparison.
For Grouped-GEMM, we port the CCQ algorithm to the vLLM~\cite{kwon2023efficient} wna16 operator, a widely adopted Group-GEMM kernel for MoE inference. The baselines (W8A16, W4A16) are experimented on the vLLM library.
%
The key distinction of CCQ lies in its use of two separate dequantization operations per computation block, as well as distinct bit shift configurations.

As shown in Table~\ref{tab:operator-performance} and Table~\ref{tab:operator-group-gemm-performance}, our CCQ-based implementation consistently delivers superior inference efficiency compared to both traditional and state-of-the-art quantization baselines. 
In GEMV scenarios, CCQ achieves up to a 2$\times$ speedup over VPTQ and closely matches the performance of standard WINT2, demonstrating that our codebook-free, bit-shift-based dequantization design effectively eliminates the memory access bottlenecks inherent to vector quantization. 
In Grouped-GEMM settings, which are critical for large-scale MoE inference, CCQ (W2A16-CCQ) not only outperforms widely adopted low-bit operators such as W8A16 and W4A16, but also maintains its latency advantage as batch size and matrix dimensions increase. 
This scalability underscores the practical value of CCQ for real-world LLM deployment, where both memory footprint and inference speed are crucial. 
Overall, the results validate that CCQ's hardware-friendly encoding and decoding pipeline enables highly efficient, large-scale inference without sacrificing model accuracy, making it a compelling solution for low-bit LLM quantization.
\begin{table}[h]
\renewcommand{\arraystretch}{1.2}
\centering
\caption{Operator-level performance comparison on GEMV kernels (execution time in milliseconds).}
\label{tab:operator-performance}
\resizebox{\linewidth}{!}{
\begin{tabular}{@{}cccc@{}}
\toprule
Matrix Shape & \multirow{2}{*}{VPTQ} & \multirow{2}{*}{WINT2-Standard} & \multirow{2}{*}{CCQ}  \\
Input Size $\times$ Output Size &  & & \\
\midrule
$4096 \times 4096$ (M=1) & 0.058 & 0.027 & 0.034  \\
$4096 \times 1024$ (M=1) & 0.033 & 0.015 & 0.017  \\
$8192 \times 8192$ (M=1) & 0.174 & 0.073 & 0.090  \\
$8192 \times 1024$ (M=1) & 0.061 & 0.019 & 0.022  \\
\midrule
$4096 \times 4096$ (M=4) & 0.205 & 0.028 & 0.035  \\
$4096 \times 1024$ (M=4) & 0.112 & 0.015 & 0.017  \\
$8192 \times 8192$ (M=4) & 0.641 & 0.076 & 0.104  \\
$8192 \times 1024$ (M=4) & 0.216 & 0.019 & 0.024  \\
\bottomrule
\end{tabular}
}
\end{table}

\begin{table}[h]
\renewcommand{\arraystretch}{1.2}
\centering
\caption{Operator-level performance comparison on Grouped-GEMM kernels (execution time in milliseconds).}
\label{tab:operator-group-gemm-performance}
\resizebox{\linewidth}{!}{
\begin{tabular}{@{}cccc@{}}
\toprule
Matrix Shape & \multirow{2}{*}{W8A16} & \multirow{2}{*}{W4A16} & \multirow{2}{*}{W2A16-CCQ}  \\
Groups $\times$ Output Size $\times$ Input Size & & & \\
\midrule
$256 \times 7168 \times 512$ (M=1)& 0.044 & 0.030 & 0.021   \\
$256 \times 1024 \times 7168$ (M=1) & 0.100 & 0.068  & 0.058  \\
\midrule
$256 \times 7168 \times 512$ (M=4) & 0.053 & 0.035  & 0.027    \\
$256 \times 1024 \times 7168$ (M=4) & 0.315 & 0.199  & 0.166  \\
\midrule
$256 \times 7168 \times 512$ (M=16) & 0.104 & 0.071 & 0.059  \\
$256 \times 1024 \times 7168$ (M=16) & 1.052 & 0.624  & 0.527   \\
\midrule
$256 \times 7168 \times 512$ (M=64) & 0.297 & 0.193  & 0.161   \\
$256 \times 1024 \times 7168$ (M=64) & 2.655 & 1.628  & 1.386    \\
\bottomrule
\end{tabular}
}
\end{table}

\section{Conclusion}
In this study, we propose \textbf{convolutional code quantization (CCQ)} to address the inference limitations inherent in vector quantization. 
By integrating convolutional codes with scalar quantization, we construct a lookup-free encoding space that achieves a linear mapping between the codebook and weight vectors, thereby optimizing inference performance. 
Meanwhile, by maintaining high-bit quantization precision for weights, we minimize the performance degradation of the model under extremely low-bit conditions. 
Through the introduction of hybrid encoding, we can flexibly apply coding quantization to improve the compression ratio. 
Furthermore, by leveraging the characteristic of normal distribution of convolutional code values across weights, we propose Code Cluster to reduce redundancy in the encoding space. 
Ultimately, we expand the encoding configurations of CCQ, which not only enhances model accuracy but also further compresses the model size.

\section{Future Work}

In the domain of extreme low-bit quantization, we identify loss-based dynamic bit allocation as a critical research direction. 
For instance, during ERNIE 4.5 quantization, we observe that allocating only the 6th MoE layer as WINT4 (instead of CCQ) could maintain high accuracy, while the other expert layers retained CCQ with bpw equal to 2.06. 
This finding prompts us to ponder whether certain individual layers in LLMs play a pivotal role in preserving the model's capabilities. Meanwhile, our proposed algorithmic framework for encode quantization can theoretically be extended to other coding theories, demonstrating significant scalability. 
This will also be one of the key directions for our future exploration. 
While our experiments validate the efficacy on the MoE architecture, further investigation is warranted for dense models under state-of-the-art 2-bit quantization methods such as QTIP~\cite{tseng2024qtip} and VPTQ~\cite{liu2024vptq}. 
Additionally, inference kernel optimization presents substantial headroom through joint consideration of permutation strategies during both quantization and runtime execution. 
Co-designing these phases could yield latency reductions, particularly for memory-bound operations on devices.

\bibliography{custom}

\end{document}